%% file: main.tex
\documentclass[10pt,twocolumn,letterpaper]{article}

\usepackage[pagenumbers]{cvpr} 

\input{preamble}
\input{math_commands}

\definecolor{cvprblue}{rgb}{0.21,0.49,0.74}
\usepackage[pagebackref,breaklinks,colorlinks,allcolors=cvprblue]{hyperref}

\title{\ourmodelplus: Improving 3D Reconstructions with Single-Step Diffusion Models}

\author{
Jay Zhangjie Wu\textsuperscript{1,2*}\quad 
Yuxuan Zhang\textsuperscript{1*}\quad 
Haithem Turki\textsuperscript{1}\quad 
Xuanchi Ren\textsuperscript{1,3,4}\quad 
Jun Gao\textsuperscript{1,3,4} \\ 
Mike Zheng Shou\textsuperscript{2}\quad 
Sanja Fidler\textsuperscript{1,3,4}\quad 
Zan Gojcic\textsuperscript{1\textdagger}\quad 
Huan Ling\textsuperscript{1,3,4\textdagger} \\
\textsuperscript{1}NVIDIA, \textsuperscript{2}National University of Singapore, \textsuperscript{3}University of Toronto, \textsuperscript{4}Vector Institute \\ \vspace{-1.0em} \\
\small\url{https://research.nvidia.com/labs/toronto-ai/difix3d}
\vspace{-2em}
}

\begin{document}

\twocolumn[{
\maketitle
\renewcommand\twocolumn[1][]{#1}
\begin{center}
    \centering
    \includegraphics[trim={0 10 0 0},clip, width=1.0\linewidth]{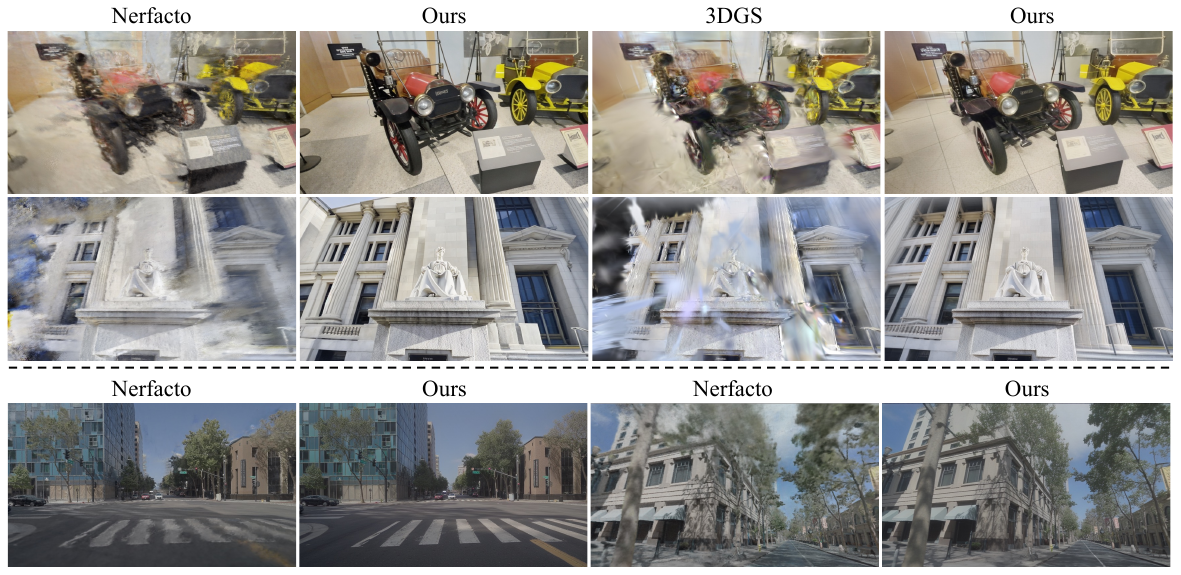}
    \vspace{-5mm}
    \captionof{figure}{\footnotesize \textbf{We demonstrate \ourmodelplus on both in-the-wild scenes (\textit{top}) and driving scenes (\textit{bottom}).} Recent Novel-View Synthesis methods struggle in sparse-input settings or when rendering views far from the input camera poses. \ourmodeldm distills the priors of 2D generative models to enhance reconstruction quality and can further act as a neural-renderer at inference time to mitigate the remaining inconsistencies. Notably, the same model effectively corrects NeRF~\cite{mildenhall2020nerf} and 3DGS~\cite{kerbl3Dgaussians} artifacts. 
    }
    \label{fig:teaser}
\end{center}
}]

\newcommand\blfootnote[1]{%
  \begingroup
  \renewcommand\thefootnote{}\footnote{#1}%
  \addtocounter{footnote}{-1}%
  \endgroup
}
\blfootnote{\textsuperscript{*,\textdagger} Equal Contribution.}

\input{sec/0_abstract}    
\input{sec/1_intro}
\input{sec/2_related_work}
\input{sec/3_background}
\input{sec/4_method}

\input{sec/5_experiments}
\input{sec/6_conclusion}
{
    \small
    \bibliographystyle{ieeenat_fullname}
    \bibliography{main}
}

\input{sec/X_suppl}

\end{document}

%% file: preamble.tex
\usepackage{svg}
\usepackage{arydshln}
\usepackage{bm}
\usepackage[linesnumbered,ruled,vlined]{algorithm2e}

\newcommand{\ourmodeldm}{\textsc{Difix}\xspace}
\newcommand{\ourmodel}{\textsc{Difix3D}\xspace}
\newcommand{\ourmodelplus}{\textsc{Difix3D+}\xspace}
\newcommand{\parahead}[1]{\vspace{1mm}\noindent\textbf{{#1}.}\ }

%% file: math_commands.tex
\usepackage{amsmath,amsfonts,bm,mathtools}

\def\eqref#1{equation~\ref{#1}}

\def\1{\bm{1}}

\def\rvepsilon{{\boldsymbol{\epsilon}}}

\def\rvc{{\mathbf{c}}}

\def\rvx{{\mathbf{x}}}
\def\rvy{{\mathbf{y}}}
\def\rvz{{\mathbf{z}}}

\def\vc{{\bm{c}}}
\def\vd{{\bm{d}}}

\def\vo{{\bm{o}}}

\def\vr{{\bm{r}}}
\def\vs{{\bm{s}}}

\def\mI{{\bm{I}}}

\DeclareMathAlphabet{\mathsfit}{\encodingdefault}{\sfdefault}{m}{sl}
\SetMathAlphabet{\mathsfit}{bold}{\encodingdefault}{\sfdefault}{bx}{n}

\def\gN{{\mathcal{N}}}

\def\@onedot{\ifx\@let@token.\else.\null\fi\xspace}

\newcommand{\E}{\mathbb{E}}

\newcommand{\R}{\mathbb{R}}



%% file: sec/0_abstract.tex
\begin{abstract}
Neural Radiance Fields and 3D Gaussian Splatting have revolutionized 3D reconstruction and novel-view synthesis task. However, achieving photorealistic rendering from extreme novel viewpoints remains challenging, as artifacts persist across representations. In this work, we introduce \ourmodelplus, a novel pipeline designed to enhance 3D reconstruction and novel-view synthesis through single-step diffusion models. At the core of our approach is \ourmodeldm, a single-step image diffusion model trained to enhance and remove artifacts in rendered novel views caused by underconstrained regions of the 3D representation.
\ourmodeldm serves two critical roles in our pipeline. First, it is used during the reconstruction phase to clean up pseudo-training views that are rendered from the reconstruction and then distilled back into 3D. This greatly enhances underconstrained regions and improves the overall 3D representation quality. More importantly, \ourmodeldm also acts as a neural enhancer during inference, effectively removing residual artifacts arising from imperfect 3D supervision and the limited capacity of current reconstruction models. \ourmodelplus is a general solution, a single model compatible with both NeRF and 3DGS representations, and it achieves an average 2$\times$ improvement in FID score over baselines while maintaining 3D consistency.
\end{abstract}

%% file: sec/1_intro.tex
\section{Introduction}
\label{sec:intro}
\begin{figure*}[t!]
   \centering
   \includegraphics[width=1.\linewidth]{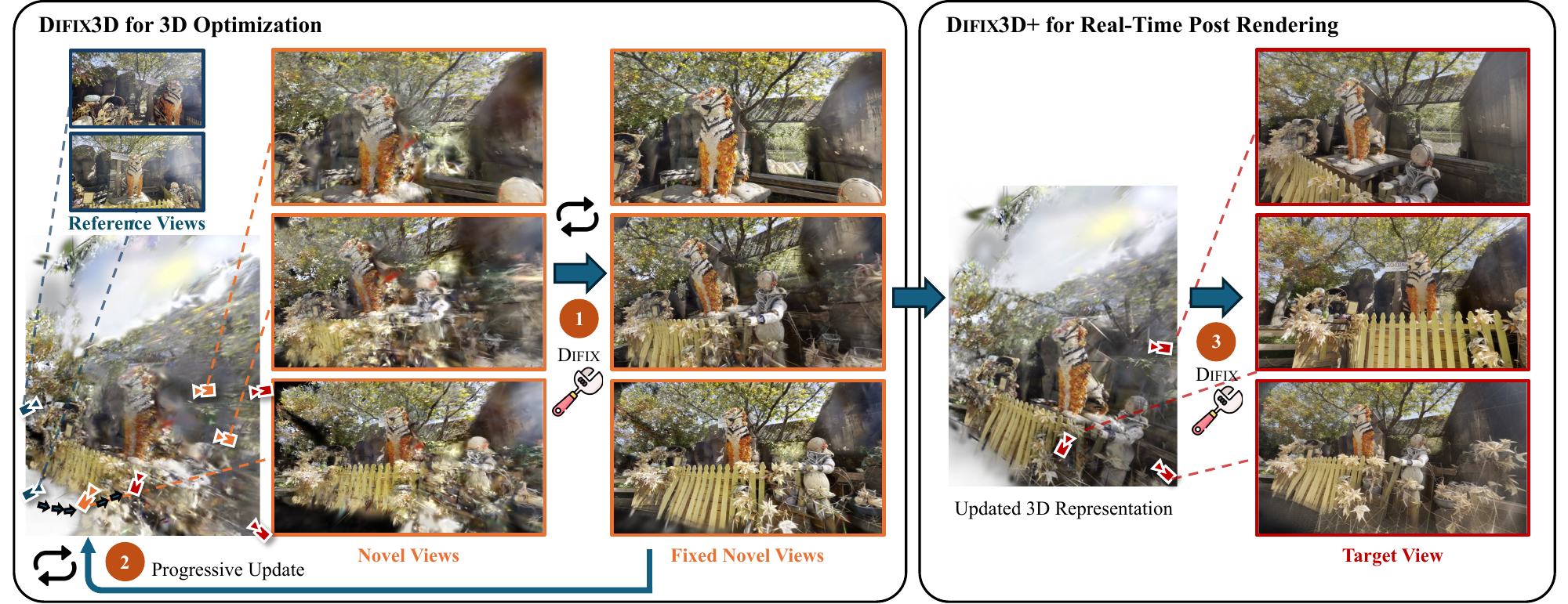}\\
\small{\textbf{\textcolor{Blue}{Blue Cameras}}: Training Views; \textbf{\textcolor{Red}{Red Cameras}}: Target Views; \\ \textbf{\textcolor{Orange}{Orange Cameras}}: Intermediate Novel views along the progressive 3D updating trajectory (\cref{sec:3d_consistency}).}
\vspace{-3mm}
   \caption{\footnotesize  \textbf{\ourmodelplus{} pipeline.}  The overall pipeline of the \ourmodelplus{} model involves the following stages:  \textbf{Step 1}: Given a pretrained 3D representation, we render novel views and feed them to \ourmodeldm  which acts as a neural enhancer, removing the artifacts and improving the quality of the noisy rendered views (\cref{sec:single-step-diffusion}). The camera poses selected to render the novel views are obtained through pose interpolation, gradually approaching the target poses from the reference ones. \textbf{Step 2}: The cleaned novel views are distilled back to the 3D representation to improve its quality (\cref{sec:3d_consistency}). Steps 1 and 2 are applied in several iterations to progressively grow the spatial extent of the reconstruction and hence ensure strong conditioning of the diffusion model (\ourmodel). \textbf{Step 3}: \ourmodeldm additional acts as a real-time neural enhancer, further improving the quality of the rendered novel views.}
   \label{fig:pipeline}
   \vspace{-5mm}
\end{figure*}
Recent advances in neural rendering, particularly Neural Radiance Fields (NeRF)~\cite{mildenhall2020nerf} and 3D Gaussian Splatting (3DGS)~\cite{kerbl3Dgaussians}, represent an important step towards photorealistic novel-view synthesis. However, despite their impressive performance near training camera views, these methods still suffer from artifacts such as spurious geometry and missing regions, especially when rendering less observed areas or more extreme novel views. The issue persists even for densely sampled captures collected under varying lighting conditions or with imperfect camera poses and calibration, hampering their suitability to real-world settings.

A core limitation of most NeRF and 3DGS approaches is their per-scene optimization framework, which requires carefully curated, view-consistent input data, and makes them susceptible to the \textit{shape-radiance ambiguity}~\cite{zhang2020nerf++}, where training images can be perfectly regenerated from a 3D representation that does not necessarily respect the underlying geometry of the scene. Without the data priors, these methods are also fundamentally limited in their ability to hallucinate plausible geometry and appearance in the underconstrained regions, and can only rely on the inherent smoothness of the underlying representation.

Unlike per-scene optimization based methods, large 2D generative models (e.g. diffusion models) are trained on internet-scale datasets, effectively learning the distribution of real-world images. Priors learned by these models generalize well to a wide range of scenes and use cases, and have been demonstrated to work on tasks such as inpainting~\cite{Grechka_2024, wang2023imagen, zhang2023towards} and outpainting~\cite{chen2024follow, wang2024your, yang2024vipversatileimageoutpainting}. However, the best way to lift these 2D priors to 3D remains unclear. Many contemporary methods query the diffusion model at each training step ~\cite{zhou2023sparsefusion, liu2023zero, poole2023dreamfusion, wu2024reconfusion}. These approaches primarily focus on optimizing object-centric scenes and scale poorly to larger environments with more expansive sets of possible camera trajectories~\cite{zhou2023sparsefusion, liu2023zero, poole2023dreamfusion}. Additionally, they are often time-consuming~\cite{wu2024reconfusion}.

In this work, we tackle the challenge of using 2D diffusion priors to improve 3D reconstruction of large scenes in an efficient manner. To this end, we build upon recent advances in single-step diffusion~\cite{sauer2025adversarial, luo2023latent, lin2024sdxl, yin2024onestep, yin2024improved}, which greatly accelerate the inference speed of text-to-image generation. We show that these single-step models retain visual knowledge that can, with minimal fine-tuning, be adapted to ``fix" artifacts present in NeRF/3DGS renderings. We use this fine-tuned model (\ourmodeldm) during the reconstruction phase to generate pseudo-training views, which when distilled back into 3D, greatly enhance quality in underconstrained regions. Moreover, as the inference speed of these models is fast, we also directly apply \ourmodeldm to the outputs of the improved reconstruction to further improve quality as a real-time post-processing step (\ourmodelplus{}). 

We make the following contributions: \textbf{(i)} We show how to adapt 2D diffusion models to remove artifacts resulting from rendering a 3D neural representation, with minimal effort. The fine-tuning process takes only a few hours on a single consumer graphics card. Despite the short training time, the same model is powerful enough to remove artifacts in rendered images from both implicit representations such as NeRF and explicit representations like 3DGS. \textbf{(ii)} We propose an update pipeline that progressively refines the 3D representation by distilling back the improved novel views, thus ensuring multi-view consistency and significantly enhanced quality of the 3D representation. Compared to contemporary methods~\cite{wu2024reconfusion, liu2023zero123} that query a diffusion model at each training time step, our approach is $>$10$\times$ faster. \textbf{(iii)} We demonstrate how single-step diffusion models enable near real-time post-processing that further improves novel view synthesis quality. \textbf{(iv)} We evaluate our approach across different datasets and present SoTA results, improving PSNR by $>$1dB and FID by $>$2$\times$ on average. 

%% file: sec/2_related_work.tex
\section{Related Work}
\label{sec:formatting}

The field of scene reconstruction and novel-view synthesis was revolutionized by the seminal NeRF~\cite{mildenhall2020nerf} and 3DGS~\cite{kerbl3Dgaussians} works, which inspired a vast corpus of follow-up efforts. In the following, we discuss a non-exhaustive list of these approaches along axes relevant to our work.
\vspace{-3mm}
\paragraph{Improving 3D reconstruction discrepancies.} Most 3D reconstruction methods assume perfect input data, yet real-world captures often include slight inconsistencies that lead to artifacts and blurriness when distilled into a 3D representation. To address this, several methods improve NeRF's robustness to noisy camera inputs by optimizng camera poses~\cite{wang2021nerf, lin2021barf, park2023camp, chen2023dbarf, meuleman2023progressively, truong2023sparf}. Other works focus on addressing lighting variations across images~\cite{martin2021nerf, VRNeRF, turki2023suds} and mitigating transient occlusions~\cite{sabour2023robustnerf}. While these methods compensate for input data inconsistencies during training, they do not entirely eliminate them. This motivates our choice to apply our fixer also at render time, further improving quality in areas affected by these discrepancies (\cref{sec:3d_consistency}).
\vspace{-3mm}
\paragraph{Priors for novel view synthesis.} Numerous works address the limitations of NeRF and 3DGS in reconstructing under-observed scene regions. Geometric priors, introduced through regularization~\cite{Niemeyer2021Regnerf, Yang2023FreeNeRF, somraj2023simplenerf} or pretrained models that provide depth~\cite{deng2022depth, roessle2022dense, wang2023sparsenerf, zhu2023FSGS} and normal~\cite{yu2022monosdf} supervision, improve rendering quality in sparse-view settings. However, these methods are sensitive to noise, difficult to balance with data terms, and yield only marginal improvements in denser captures. Other works train feed-forward neural networks with posed multi-view data collected across numerous scenes. At render time, these approaches aggregate information from neighboring reference views to either enhance a previously rendered view~\cite{zhou2023nerflix} or directly predict a novel view~\cite{yu2020pixelnerf, mvsnerf, ren2024scube, lu2024infinicube}. While these deterministic methods perform well near reference views, they often produce blurry results in ambiguous regions where the distribution of possible renderings is inherently multi-modal.
\vspace{-3mm}
\paragraph{Generative priors for novel view synthesis.} Recently, priors learned by the generative models have been increasingly used to enhance novel view synthesis. GANeRF~\cite{roessle2023ganerf} trains a per-scene generative adversarial network (GAN) that enhances NeRF's realism. Many other works use diffusion models that learn strong and generalizable priors from internet scale datasets. These diffusion models can either directly generate novel views with minimal fine-tuning~\cite{guo2023animatediff, yu2024viewcrafter, gao2024cat3d, zhang2024recapture} or guide the optimization of a 3D representation. In the latter case, the diffusion model often serves as \emph{scorer} that need to be queried during each optimization step~\cite{gu2023nerfdiff, wu2024reconfusion, zhou2023sparsefusion, liu2023zero, warburg2023nerfbusters}, which significantly slows down training. In contrast, Deceptive-NeRF~\cite{liu2023deceptive} and, concurrently with our work, 3DGS-Enhancer~\cite{liu20243dgs} use diffusion priors to enhance pseudo-observations rendered from the 3D representation, augmenting the training image set for fine-tuning the 3D representation. Since this approach avoids querying the diffusion model at every training step, the overhead is significantly reduced. While our work follows a similar direction, we diverge in two key aspects: (i) we introduce a progressive 3D update pipeline that effectively corrects artifacts even in extreme novel views while preserving long-range consistency and (ii) we use our model both during optimization and at render-time, leading to improved visual quality.

%% file: sec/3_background.tex
\section{Background}
\label{sec:background}
\paragraph{3D Scene Reconstruction and Novel-View Synthesis.} 
Neural Radiance Fields (NeRFs) have transformed the field of novel-view synthesis by modeling scenes as an emissive volume encoded within the weights of a coordinate-based multilayer perceptron (MLP).
This MLP can be queried at any spatial location to return the view-dependent radiance $\vc \in \mathbb{R}^3$ and volume density $\sigma \in \mathbb{R}$. The color of a ray $\mathbf{r}(\tau) = \vo + t\vd$ with origin $\vo \in \mathbb{R}^3$ and direction $\vd \in \mathbb{R}^3$ can then be rendered from the above representation by sampling points along the ray and accumulating their radiance through volume rendering as:
\begin{align}
    \label{eq:volume_rendering}
    \mathcal{C}(\mathbf{p}) & = \sum_{i=1}^N \alpha_i \mathbf{c}_i  \prod_j^{i-1} (1 - \alpha_i)
\end{align}
where $\alpha_i = (1 - \exp(-\alpha_i \delta_i))$, $N$ denotes the number of samples along the ray, and $\delta_i$ is the step size used for quadrature.  

Instead of representing scenes as a continuous neural field, 3D Gaussian Splatting~\cite{kerbl3Dgaussians} uses volumetric particles parameterized by their positions $\boldsymbol{\mu} \in \mathbb{R}^3$, rotation $\mathbf{r} \in \mathbb{R}^4$, scale  $\mathbf{s} \in \mathbb{R}^3$, opacity $\eta \in \mathbb{R}$ and color $\mathbf{c}_i$. Novel views can be rendered from this representation using the same volume rendering formulation from \cref{eq:volume_rendering}, where
\begin{align}
    \alpha_i &= \eta_i \,\textrm{exp}\left[-\frac{1}{2}\left(\mathbf{p}-{\boldsymbol{\mu}}_i\right)^\top{\bm{\Sigma}}^{-1}_i\left(\mathbf{p}-{\boldsymbol{\mu}}_i\right)\right]
\end{align}
with $\bm{\Sigma} =\bm{R}\bm{S}\bm{S}^{T}\bm{R}^{T}$ and $\mathbf{R} \in \text{SO}(3)$ and $\mathbf{S} \in \mathbb{R}^{3 \times 3}$ are the matrix representation of $\vr$ and $\vs$, respectively. The number $N$ of Gaussians that contribute to each pixel is determined through tile-base rasterization.

\paragraph{Diffusion Models.} DMs~\cite{sohl2015deep,ho2020ddpm,song2020score} learn to model the data distribution $p_{\text{data}}(\rvx)$ through \emph{iterative denoising} and are trained with \textit{denoising score matching}~\cite{hyvarinen2005scorematching,lyu2009scorematching,vincent2011,sohl2015deep,song2019generative,ho2020ddpm,song2020score}. Specifically, to train a diffusion model, \emph{diffused} versions $\rvx_\tau = \alpha_\tau \rvx + \sigma_\tau \rvepsilon$ of the data $\rvx \sim p_{\text{data}}$ are generated, by progressively adding Gaussian noise $\rvepsilon \sim \gN(\mathbf{0}, \mI)$. 
Learnable parameters $\bm{\theta}$ of the denoiser model $\mathbf{F}_{\bm{\theta}}$ are optimized using the denoising score matching objective: 
\begin{align}
\E_{\rvx \sim p_{\text{data}}, \tau \sim p_{\tau}, \rvepsilon \sim \gN(\mathbf{0}, \mI)} \left[\Vert \rvy - \mathbf{F}_\theta(\rvx_\tau; \rvc, \tau) \Vert_2^2 \right],
\label{eq:diffusionobjective}
\end{align}
where $\rvc$ represents optional conditioning information, such as a text prompt or image context. Depending on the model formulation, the target vector $\rvy$ is usually set as the added noise $\rvepsilon$.
Finally, $p_\tau$ denotes a uniform distribution over the diffusion time variable $\tau$. 
In practice a fixed discretization can be used~\cite{ho2020ddpm}. In this setting, $p_\tau$ is often chosen as a uniform distribution, $p_\tau \sim \mathcal{U}(0, 1000)$.
The maximum diffusion time $\tau = 1000$ is generally set such that the input data is fully transformed into Gaussian noise. 

%% file: sec/4_method.tex
\section{Boosting 3D Reconstruction with DM priors}

Given a collection of RGB images and corresponding camera poses, our goal is to reconstruct a 3D representation that enables realistic novel view synthesis from arbitrary viewpoints, with particular emphasis on underconstrained regions distant from the input camera positions. To achieve this, we leverage the strong generative priors of a pre-trained diffusion model during: \textbf{(i) optimization} to iteratively augment the training set with clean pseudo-views that improve the underlying 3D representation in distant and unobserved areas, and \textbf{(ii) inference} as a real-time post-processing step that further reduces artifacts caused by insufficient or inconsistent training supervision.

We first describe how to adapt a pretrained diffusion model into an image-to-image translation model that removes artifacts present in neural rendering methods (\cref{sec:single-step-diffusion}) and the data curation strategy used to fine-tune this model (\cref{sec:data-curation}). We then show how to use our fine-tuned diffusion model to improve the novel view synthesis quality of 3D representations in \cref{sec:3d_consistency}.  

\begin{figure*}[t!]
    \centering
    \vspace{-6mm}
    \includegraphics[width=0.9\linewidth]{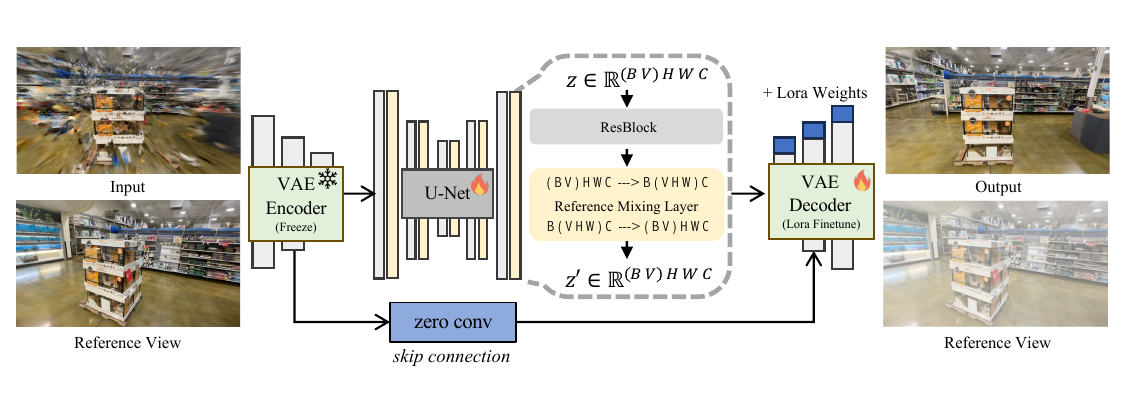}
    \vspace{-9mm}
    \caption{\footnotesize \textbf{\ourmodeldm architecture.} \ourmodeldm takes a noisy rendered image and a reference views as input (\emph{left}), and outputs an enhanced version of the input image with reduced artifacts (\emph{right}). \ourmodeldm also generates identical reference views, which we discard in practice and hence depict transparent. The model architecture consists of a U-Net structure with a cross-view reference mixing layer (\cref{sec:single-step-diffusion}) to maintain consistency across reference views. \ourmodeldm is fine-tuned from SD-Turbo, using a frozen VAE encoder and a LoRA fine-tuned decoder.}
    \label{fig:architecture}
    \vspace{-4mm}
\end{figure*}

We visualize the overall \ourmodelplus pipeline in \cref{fig:pipeline} and the architecture of our \ourmodeldm diffusion model in \cref{fig:architecture}.

\subsection{{\large\ourmodeldm}: From a pretrained diffusion model to a 3D Artifact Fixer}
\label{sec:single-step-diffusion}

Given a rendered novel view $\Tilde{I}$ that may contain artifacts from the 3D representation and a set of clean reference views $I_{\text{ref}}$, our model produces a refined novel view prediction $\hat{I}$. We build our model on top of a single-step diffusion model SD-Turbo~\cite{sauer2025adversarial}, which has proven effective for image-to-image translation tasks~\cite{parmar2024one}, for efficiency reasons and to enable real-time post-processing during inference.

\parahead{Reference view conditioning} We condition our model on a set of clean reference views $I_{\text{ref}}$, which in practice, we select as the closest training view.
Inspired by video ~\cite{blattmann2023videoldm,singer2023makeavideo,ho2022imagen,zhou2023magicvideo,wang2023videofactory,wang2023lavie,zhang2024show,ge2022pyoco,wu2023tune,guo2023animatediff,girdhar2023emu,agarwal2025cosmos} and multi-view diffusion models~\cite{liu2023zero123,qian2023magic123,liu2023syncdreamer,shi2023mvdream,shi2023zero123plus,liu2023one2345,xu2023dmv3d,long2023wonder3d}, we adapt the self-attention layers into a \emph{reference mixing layer} to capture cross-view dependencies. We start from concatenating novel view $\Tilde{I}$ and reference views $I_{\text{ref}}$ on an additional view dimension and frame-wise encoded into latent space $\mathcal{E}( (\Tilde{I}, I_{\text{ref}}) )=\rvz \in \R^{V \times C \times H \times W}$,  where $C$ is the number of latent channels, $V$ is input number of views (reference views and target views) and $H$ and $W$ are the spatial latent dimensions. The \emph{reference mixing layer} operates by first shifting the view axis to the spatial axis and reshaping back after the self-attention operation as follows (using \texttt{einops}~\cite{rogozhnikov2022einops} notation): 
\vspace{-1mm}
\begin{align*}
\mathbf{z}' &\leftarrow \texttt{rearrange}(\mathbf{z}, \; \texttt{b c v (hw)} \rightarrow \texttt{b c (vhw)}) \\
\mathbf{z}' &\leftarrow l_\phi^i(\mathbf{z}', \mathbf{z}') \\
\mathbf{z}' &\leftarrow \texttt{rearrange}(\mathbf{z}', \; \texttt{b c (vhw)} \rightarrow \texttt{b c v (hw)}) ,
\end{align*}
where $l_\phi^i$ is a self-attention layer applied over the \texttt{vhw} dimension. This design allows us to inherit all module weights from the original 2D self-attention. We found this adaptation effective for capturing key information (e.g., objects, color, texture) from reference views, especially when the quality of the original novel view is severely degraded. 

\begin{figure}[t]
    \centering
    \includegraphics[width=0.9\linewidth]{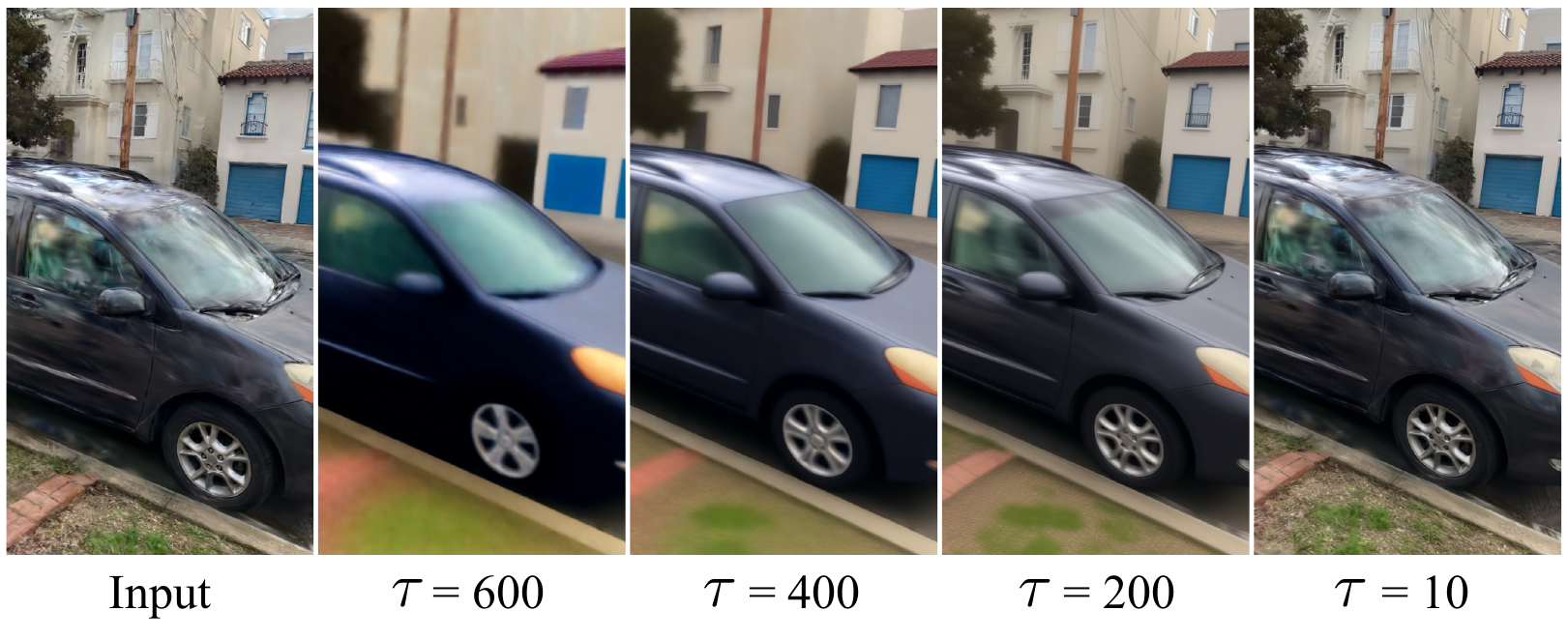}
    \resizebox{\linewidth}{!}{
    \begin{tabular}{@{}l|cccccc@{}}
    \hline
    \textbf{$\tau$} & 1000     & 800     & 600     & 400     & 200     & 10       \\
    \hline
    PSNR     & 12.18 & 13.63 & 15.64 & 17.05 & \textbf{17.73} & 17.72 \\
    SSIM     & 0.4521  & 0.5263  & 0.6129  & 0.6618  & \textbf{0.6814}  & 0.6752  \\
    \hline
    \end{tabular}
    }
    \vspace{-2mm}
    \caption{\footnotesize \textbf{Noise level.} To validate our hypothesis that the distribution of images with NeRF/3DGS artifacts is similar to the distribution of noisy images used to train SD-Turbo~\cite{sauer2025adversarial}, we perform single-step ``denoising" at varying noise levels. At higher noise levels (e.g., $\tau = 600$), the model effectively removes artifacts but also alters the image context. At lower noise levels (e.g., $\tau = 10$), the model makes only minor adjustments, leaving most artifacts intact. $\tau = 200$ strikes a good balance, removing artifacts while preserving context, and achieves the highest metrics.}
    \vspace{-5mm}
    \label{tab:psnr_ssim}
\end{figure}

\parahead{Fine-tuning} We fine-tune SD-Turbo~\cite{sauer2025adversarial} in a similar manner to Pix2pix-Turbo~\cite{parmar2024one}, using a frozen VAE encoder and a LoRA fine-tuned decoder. As in Image2Image-Turbo~\cite{parmar2024one}, we train our model to directly take the degraded rendered image $\Tilde{I}$ as input, rather than random Gaussian noise, but apply a lower noise level ($\tau=200$ instead of $\tau=1000$). Our key insight is that the distribution of images degraded by neural rendering artifacts $\Tilde{I}$ resembles the distribution of images $\rvx_\tau$ originally used to train the diffusion model at a specific noise level $\tau$ (\cref{sec:background}). We validate this intuition by performing single-step ``denoising" of rendered NeRF/3DGS images with artifacts, using a pre-trained SD-Turbo model. As shown in \cref{tab:psnr_ssim}, $\tau = 200$ achieves the best results both visually and in terms of metrics.

\vspace{-4mm}
\paragraph{Losses.}

We supervise our diffusion model with losses derived from readily available 2D supervision. 
We use the L2 difference between the model output $\hat{I}$ and the ground-truth image $I$ along with a perceptual LPIPS loss (as described in the supplement) in addition to a style loss term which encourages sharper details. We do so via a Gram matrix loss that 
defined as the L2 norm of the auto-correlation of VGG-16 features~\cite{reda2022film}:
\begin{equation}
\mathcal{L}_{\text{Gram}} = \frac{1}{L} \sum_{l=1}^L \beta_l \left\| G_l(\hat{I}) - G_l(I) \right\|_2,
\end{equation}
\noindent with the Gram matrix at layer $l$ defined as:
\begin{equation}
G_l(I) = \phi_l(I)^\top \phi_l(I).
\end{equation}
The final loss used to train our model is the weighted sum of the above terms: $\mathcal{L} = \mathcal{L}_{\text{Recon}} + \mathcal{L}_{\text{LPIPS}} + 0.5\mathcal{L}_{\text{Gram}}$.

\subsubsection{Data Curation}
\label{sec:data-curation}

To supervise our model with the above loss terms, we require access to a large dataset consisting of pairs of images containing artifacts typical in novel-view synthesis and the corresponding ``clean" ground truth images. A seemingly straightforward strategy would be to train a 3D representation with every $n$th frame and pair the remaining ground truth images with the rendered ``novel" views. This \textbf{sparse reconstruction} strategy works well on the DL3DV dataset \cite{ling2024dl3dv}, which contains camera trajectories that allow us to sample novel views with significant deviation. 
However, it is suboptimal in most other novel view synthesis datasets~\cite{mildenhall2019llff, barron2022mipnerf360}
where even held-out views largely observe the same region as the training views~\cite{warburg2023nerfbusters}. We therefore explore various strategies to increase the amount of training examples (\cref{tab:data_curation}) :

\begin{table}[t]
    \centering
    \resizebox{\linewidth}{!}{
    \begin{tabular}{@{}l|cccc@{}}
    \hline
    & Sparse & Cycle & Cross & Model \\
    & Reconstruction & Reconstruction & Reference & Underfitting \\
    \hline
    DL3DV \cite{ling2024dl3dv} & \checkmark &  &  & \checkmark \\
    Internal RDS &   & \checkmark & \checkmark & \checkmark \\
    \hline
    \end{tabular}
    }
    \vspace{-2mm}
    \caption{\footnotesize \textbf{Data curation.} We curate a paired dataset featuring common artifacts in novel-view synthesis. For DL3DV scenes  \cite{ling2024dl3dv}, we employ sparse reconstruction and model underfitting, while for internal real driving scene (RDS) data, we utilize cycle reconstruction, cross reference, and model underfitting techniques.}
    \label{tab:data_curation}
    \vspace{-5mm}
\end{table}

\textbf{Cycle Reconstruction.} In nearly linear trajectories, such as those found in autonomous driving datasets, we first train a NeRF on the original path, and then render views from a trajectory shifted 1-6 meters horizontally (which we found to work well empirically). We then train a second NeRF representation against these rendered views and use this second NeRF to render degraded views for the original camera trajectory (for which we have ground truth).
\begin{figure*}[t]
  \centering
    \includegraphics[width=\linewidth]{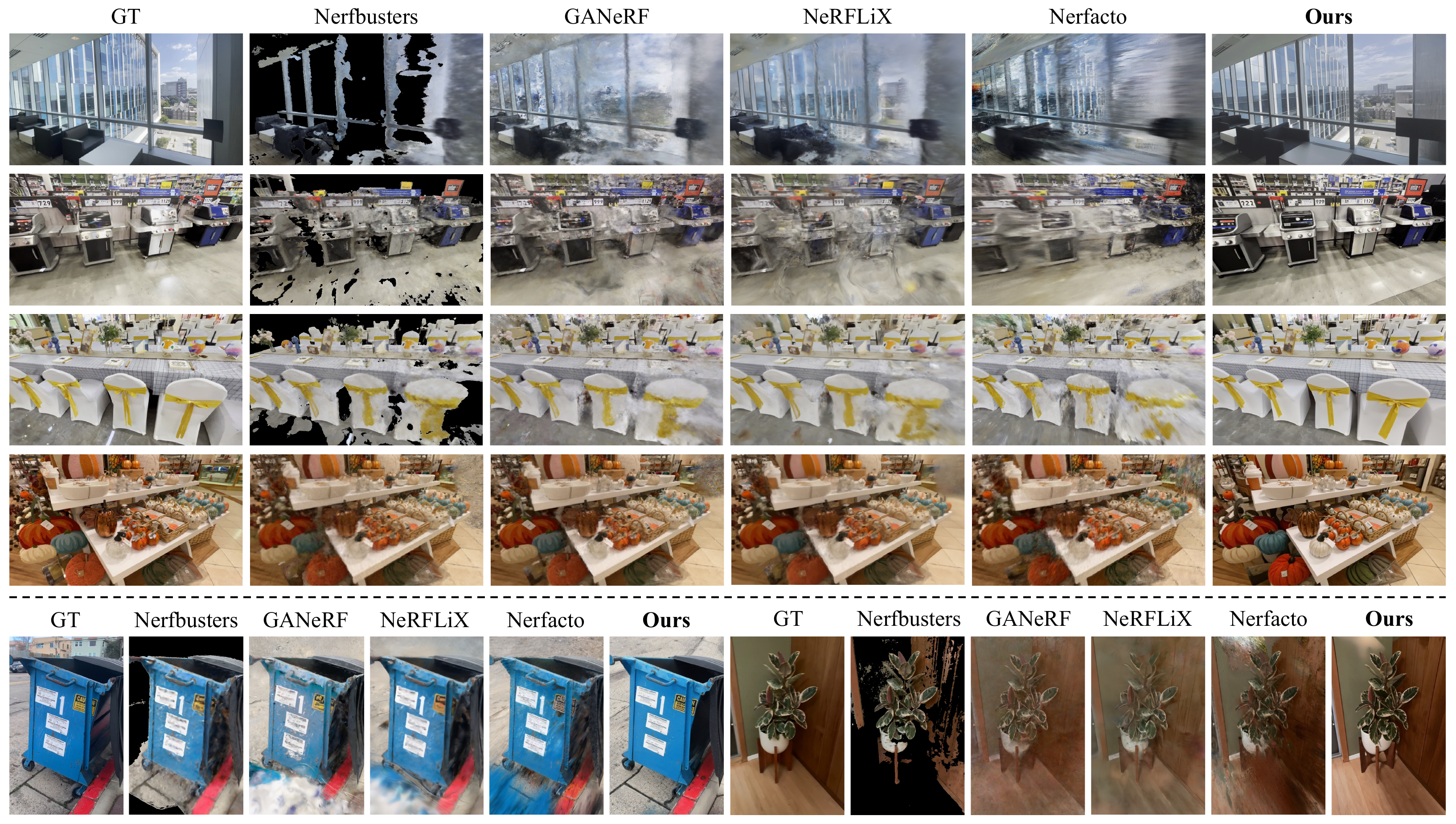}
    \vspace{-7mm}
    \caption{\footnotesize \textbf{In-the-wild artifact removal.} We show comparisons on held-out scenes from the DL3DV dataset~\cite{ling2024dl3dv} (\textit{top}, above the dashed line) and the Nerfbusters~\cite{warburg2023nerfbusters} dataset (\textit{bottom}). \ourmodelplus corrects significantly more artifacts that other methods.} 
    \label{fig:qualitative_comparision}
    \vspace{-4mm}
\end{figure*}

\textbf{Model Underfitting.} To generate more salient artifacts than those obtained by merely holding out views, we underfit our reconstruction by training it with a reduced number of epochs (25\%-75\% of the original training schedule).
We then render views from this underfitted reconstruction and pair them with the corresponding ground truth images.

\textbf{Cross Reference.} For multi-camera datasets, we train the reconstruction model solely with one camera and render images from the remaining held out cameras.
We ensure visual consistency by selecting cameras with similar ISP.

\subsection{{\large \ourmodelplus}: NVS with Diffusion Priors}
\label{sec:3d_consistency}

Our trained diffusion model can be directly applied to enhance rendered novel views during inference (see (a) in \cref{tab:ablation_stages}). However, due to the generative nature of the model, this results in inconsistencies across different poses/frames, especially in under-observed and noisy regions where our model needs to hallucinate high-frequency details or even larger areas. An example is shown in \cref{fig:visual_ablation}, where the first column displays the NeRF result. Directly using \ourmodeldm to correct this novel view leads to inconsistent fixes. To address this issue, we distill the outputs of our diffusion model back into the 3D representation during training. This not only improves the multi-view consistency, but also leads to higher perceptual quality of the rendered novel views (see (b-c) in \cref{tab:ablation_stages}). Furthermore, we apply a final neural enhancer step during rendering inference, effectively removing residual artifacts.  (see (d) in \cref{tab:ablation_stages}).

\begin{table*}[t]
  \centering
  \resizebox{0.75\linewidth}{!}{
  \begin{tabular}{@{}l|cccc|cccc@{}}
    \toprule
    & \multicolumn{4}{c|}{Nerfbusters Dataset} & \multicolumn{4}{c}{DL3DV Dataset} \\
    Method & PSNR$\uparrow$ & SSIM$\uparrow$ & LPIPS$\downarrow$ & FID$\downarrow$ & PSNR$\uparrow$ & SSIM$\uparrow$ & LPIPS$\downarrow$ & FID$\downarrow$ \\
    \midrule
    Nerfbusters~\cite{warburg2023nerfbusters} & 17.72 & 0.6467 & 0.3521 & 116.83 & 17.45 & 0.6057 & 0.3702 & 96.61 \\
    GANeRF~\cite{roessle2023ganerf} & 17.42 & 0.6113 & 0.3539 & 115.60 & 17.54 & 0.6099 & 0.3420 & 81.44 \\
    NeRFLiX~\cite{zhou2023nerflix} & 17.91 & \underline{0.6560} & 0.3458 & 113.59 & 17.56 & \underline{0.6104} & 0.3588 & 80.65 \\
    Nerfacto~\cite{tancik2023nerfstudio} & 17.29 & 0.6214 & 0.4021 & 134.65 & 17.16 & 0.5805 & 0.4303 & 112.30 \\
    \ourmodel (Nerfacto) & \underline{18.08} & 0.6533 & \underline{0.3277} & \underline{63.77} & \underline{17.80} & 0.5964 & \underline{0.3271} & \underline{50.79} \\
    \ourmodelplus (Nerfacto) & \textbf{18.32} & \textbf{0.6623} & \textbf{0.2789} & \textbf{49.44} & \textbf{17.82} & \textbf{0.6127} & \textbf{0.2828} & \textbf{41.77} \\
    \hline
    3DGS~\cite{kerbl3Dgaussians} & 17.66 & 0.6780 & 0.3265 & 113.84 & 17.18 & 0.5877 & 0.3835 & 107.23 \\
    \ourmodel (3DGS) & \underline{18.14} & \underline{0.6821} & \underline{0.2836} & \underline{51.34} & \underline{17.80} & \underline{0.5983} & \underline{0.3142} & \underline{50.45} \\
    \ourmodelplus (3DGS) & \textbf{18.51} & \textbf{0.6858} & \textbf{0.2637} & \textbf{41.77} & \textbf{17.99} & \textbf{0.6015} & \textbf{0.2932} & \textbf{40.86} \\
    \bottomrule
  \end{tabular}
  }
  \vspace{-1mm}
  \caption{\footnotesize \footnotesize \textbf{Quantitative comparison on Nerfbusters and DL3DV datasets}. The best result is highlighted in \textbf{bold}, and the second-best is \underline{underlined}.}
  \vspace{-3mm}
  \label{tab:main_results}
\end{table*}

\vspace{-3mm}
\paragraph{\ourmodel: Progressive 3D updates.}
Strong conditioning of our diffusion-model on the rendered novel views and the reference views is crucial for achieving multi-view consistency and high fidelity to the input views. When the desired novel trajectory is too far from the input views, the conditioning signal becomes weaker and the diffusion model is forced to hallucinate more. We therefore adopt an iterative training scheme similar to Instruct-NeRF2NeRF~\cite{haque2023instruct} that progressively grows the set of 3D cues that can be rendered (multi-view consistently) to novel views and hence increases the conditioning for the diffusion model.

Specifically, given a set of target views, we begin by optimizing the 3D representation using the reference views. After every 1.5k iterations, we slightly perturb the ground-truth camera poses toward the target views, render the resulting novel view, and refine the rendering using the diffusion model trained in \cref{sec:single-step-diffusion}. The refined images are then added to the training set for another 1.5k iteration of training.  By progressively perturbing the camera poses, refining the novel views, and updating the training set, this approach gradually improves 3D consistency and ensures high-quality, artifact-free renderings at the target views.

This progressive process allows us to progressively increase the overlap of 3D cues between the reference and target views, ultimately achieving consistent, artifact-free renderings. \textit{See Supplementary Material for additional details about 3D update training.}

\vspace{-4mm}
\paragraph{\ourmodelplus: With Real time Post Render Processing}
Due to the slight multi-view inconsistencies of the enhanced novel views that we are distilling, and the limited capacity of reconstruction methods to represent sharp details, some regions remain blurry (the second last column in \cref{fig:visual_ablation}). To further enhance the novel views, we use our diffusion model as the final post-processing step at render time, resulting in improvement across all perceptual metrics ((d) in \cref{tab:ablation_stages}), while maintaining a high degree of consistency. Since \ourmodeldm is a single-step model, the additional rendering time is only 76 ms on an NVIDIA A100 GPU, over 10$\times$ faster than standard diffusion models with multiple denoising steps.

%% file: sec/5_experiments.tex
\section{Experiments}

We first evaluate \ourmodelplus on in-the-wild scenes against several baselines and show its ability to enhance both NeRF and 3DGS-based pipelines (\cref{sec:eval-in-the-wild}). We further evaluate the generality of our solution by enhancing automotive scenes (\cref{sec:eval-av}). We ablate our design in \cref{sec:ablations}.

\subsection{In-the-Wild Artifact Removal}
\label{sec:eval-in-the-wild}

\paragraph{\ourmodeldm training.} We train \ourmodeldm on a random selection of 80\% of scenes (112 out of a total of 140) from the DL3DV~\cite{ling2024dl3dv} benchmark dataset. We generate 80,000 noisy-clean image pairs using the dataset curation strategies listed in \cref{tab:data_curation}, and simulate NeRF and 3DGS-based artifacts in a 1:1 ratio.
\vspace{-3mm}
\paragraph{Evaluation protocol.} We evaluate \ourmodelplus with Nerfacto~\cite{tancik2023nerfstudio} and 3DGS~\cite{kerbl3Dgaussians} backbones on the 28 held out scenes from the DL3DV~\cite{ling2024dl3dv} benchmark and the 12 captures in the Nerfbusters~\cite{warburg2023nerfbusters} dataset. We partition each scene into a set of reference views used during training and evaluate on the left-out target views. We generate these splits for DL3DV by partitioning frames into two clusters based on camera position, ensuring a substantial deviation between reference and target views. We select reference and target views in the Nerfbusters dataset following their recommended protocol~\cite{warburg2023nerfbusters}.
\vspace{-3mm}
\paragraph{Baselines.} We compare our Nerfacto and 3DGS \ourmodelplus variants to their base methods. We also compare to Nerfbusters~\cite{warburg2023nerfbusters}, which uses a 3D diffusion model to remove artifacts from NeRF\footnote{Nerfbusters~\cite{warburg2023nerfbusters} uses a visibility map extracted from a NeRF model trained on a combination of training and evaluation views and remove pixels that fall outside of that visibility map. This results in missing regions in ~\cref{fig:qualitative_comparision}}, GANeRF~\cite{roessle2023ganerf}, which train per-scene GAN that is used to enhance the realism of the scene representation, and NeRFLiX~\cite{zhou2023nerflix}, which aggregates information from nearby reference views at inference time to improve novel view synthesis quality. We use the gsplat library\footnote{https://github.com/nerfstudio-project/gsplat} for 3DGS-based experiments and the official implementation for all other methods and baselines. 
\vspace{-3mm}
\paragraph{Metrics.} We calculate PSNR, SSIM~\cite{1284395}, LPIPS~\cite{johnson2016perceptual} as well as FID score ~\cite{heusel2017gans} on novel views. More details are available in the Supplementrary Material.
\vspace{-3mm}
\paragraph{Results.} We provide quantitative results in \cref{tab:main_results}. Our method outperforms all comparison methods by a significant margin across all metrics. Both \ourmodelplus variants reduce LPIPS by 0.1 and FID by almost 3$\times$ relative to their respective NeRF and 3DGS backbones, highlighting a significant improvement in perceptual quality and visual fidelity. Furthermore, \ourmodelplus also enhances PSNR, a pixel-wise metric sensitive to color shifting, by about 1db, indicating that \ourmodelplus maintains a high degree of fidelity with original views (\cref{sec:3d_consistency}). We provide qualitative examples in \cref{fig:qualitative_comparision} that show how \ourmodelplus corrects significantly more artifacts that other methods, and additional videos in the supplement to further illustrate how we maintain a high degree of consistency across rendered frames.

\subsection{Automotive Scene Enhancement}
\label{sec:eval-av}

\begin{figure}[t]
    \centering
    \includegraphics[width=\linewidth]{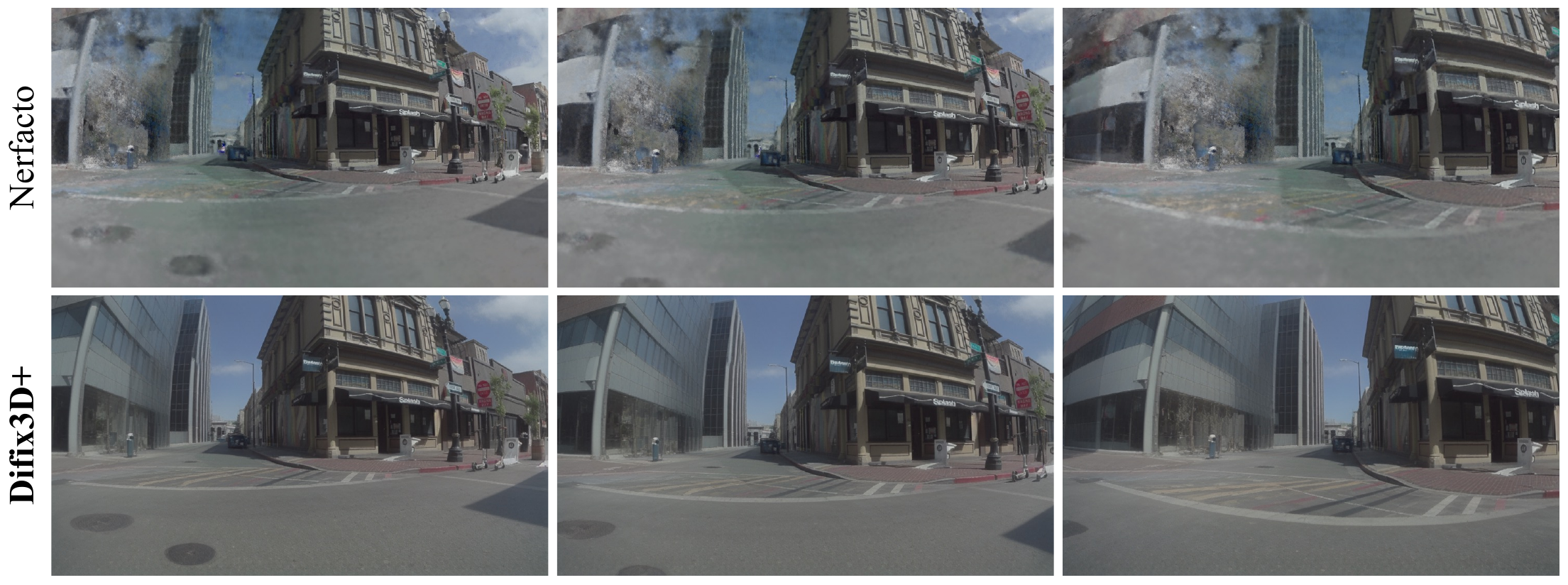}
    \vspace{-6mm}
    \caption{\footnotesize \textbf{Qualitative results on the RDS dataset.} \ourmodeldm for RDS was trained on 40 scenes and 100,000 paired data samples.}
    \label{fig:nre}
    \vspace{-2mm}
\end{figure}

\begin{table}[t!]
  \centering
  \resizebox{0.9\linewidth}{!}{
  \begin{tabular}{@{}l|cccc@{}}
    \toprule
    Method & PSNR$\uparrow$ & SSIM$\uparrow$ & LPIPS$\downarrow$ & FID$\downarrow$ \\
    \midrule
    Nerfacto & 19.95 & 0.4930 & 0.5300 & 91.38 \\
    Nerfacto + NeRFLiX & 20.44 & 0.5672 & 0.4686 &  116.28   \\
    Nerfacto + \ourmodel & 21.52 & 0.5700 & 0.4266 & 77.83  \\
    Nerfacto + \ourmodelplus & \textbf{21.75} & \textbf{0.5829} & \textbf{0.4016} & \textbf{73.08}  \\
    \bottomrule
  \end{tabular}
  }
  \vspace{-2mm}
  \caption{\footnotesize \textbf{Comparison of quantitative results on RDS dataset}. The best result is highlighted in bold.}
  \label{tab:av_all_results}
  \vspace{-3mm}
\end{table}

\begin{figure}[t]
    \centering
    \includegraphics[width=\linewidth]{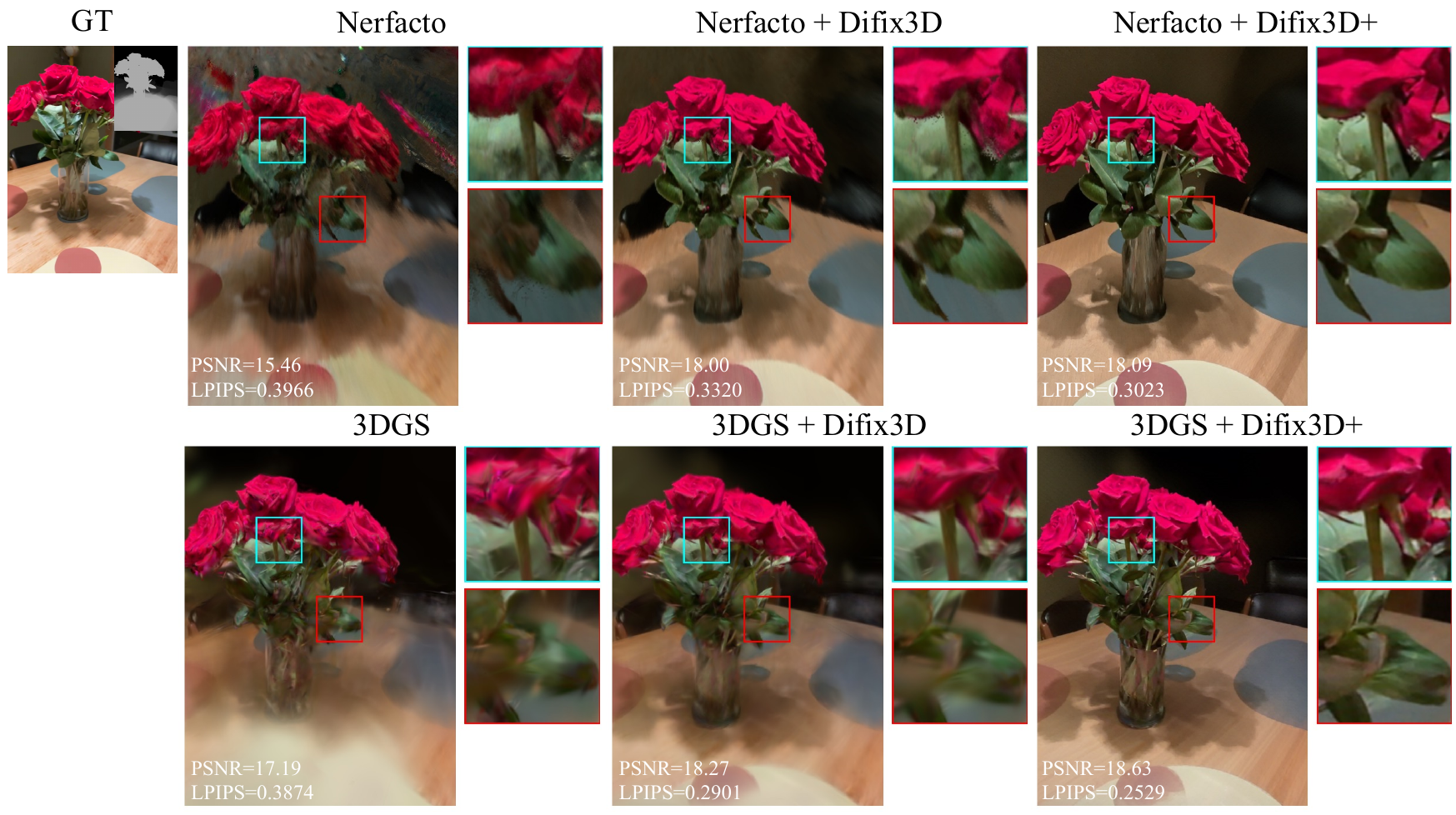}
    \vspace{-5mm}
    \caption{\footnotesize \textbf{Qualitative ablation of real-time post-render processing:} \ourmodelplus uses an additional neural enhancer step that effectively removes residual artifacts, resulting in higher PSNR and lower LPIPS scores. The images displayed in green or red boxes correspond to zoomed-in views of the bounding boxes drawn in the main images. } 
    \label{fig:compare_plus}
    \vspace{-2mm}
\end{figure}

\begin{figure}[t]
    \centering
    \includegraphics[width=\linewidth]{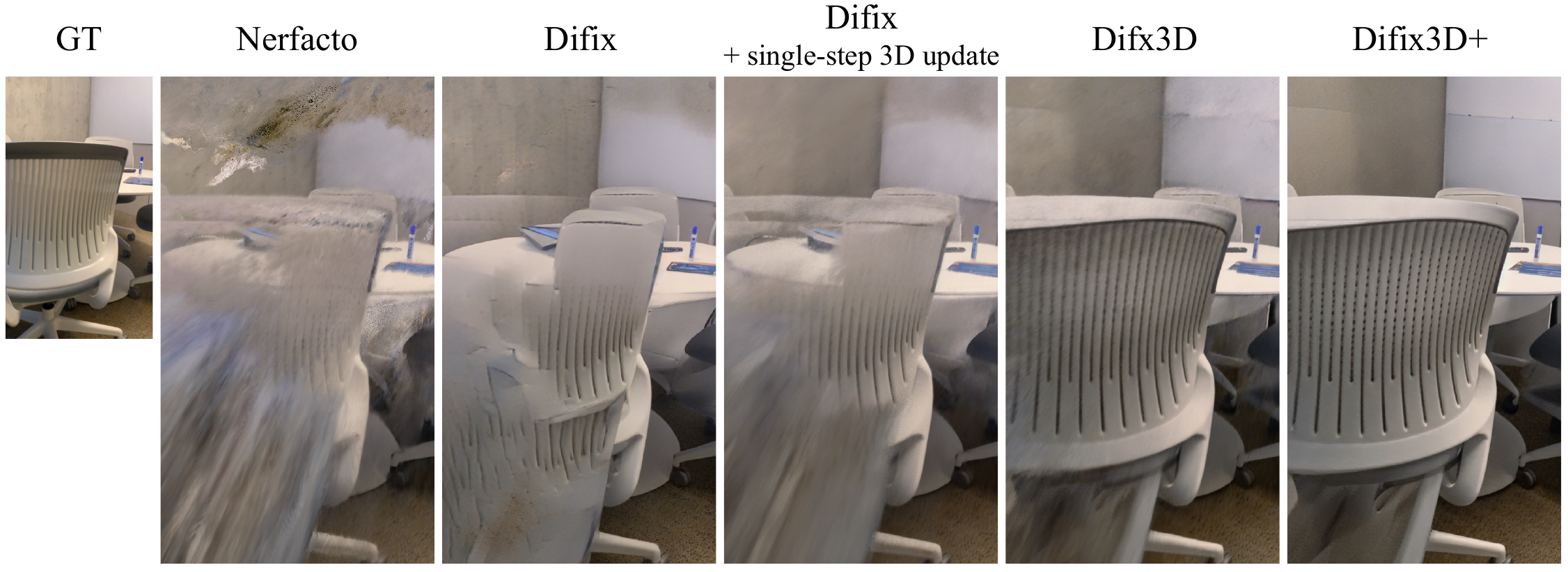}
    \caption{\footnotesize \textbf{Qualitative ablation results of \ourmodelplus}: The columns, labeled by method name, correspond to the rows in \cref{tab:ablation_stages}.}
    \label{fig:visual_ablation}
\end{figure}

\vspace{-1mm}
\paragraph{\ourmodeldm training.} We construct an in-house real driving
scene (RDS) dataset. The automotive capture rig contains three cameras with 40 degree overlaps between each camera. We train \ourmodeldm with 40 scenes and generate 100,000 image pairs using the augmentation strategies listed in \cref{tab:data_curation}.
\vspace{-3mm}
\paragraph{Evaluation protocol.} We evaluate \ourmodelplus with a Nerfacto backbone on 20 scenes (none of which are used during \ourmodeldm training). We train NeRF with the center camera and evaluate the other two cameras as novel views.
\vspace{-3mm}
\paragraph{Baselines and metrics.} We compare our method to its NeRF baseline and NeRFliX~\cite{zhou2023nerflix}. We use the same evaluation metrics as in \cref{sec:eval-in-the-wild}.

\begin{table}[t!]
  \centering
  \resizebox{\linewidth}{!}{
  \begin{tabular}{@{}l|cccc@{}}
    \toprule
    Method & PSNR$\uparrow$ & SSIM$\uparrow$ & LPIPS$\downarrow$ & FID$\downarrow$ \\
    \midrule
    Nerfacto & 17.29 & 0.6214 & 0.4021 & 134.65 \\
    + (a) \textbf{(\ourmodeldm)} & 17.40 & 0.6279 & 0.2996 & 49.87 \\
    + (a) + (b)  (\ourmodeldm + single-step
3D update)& 17.97 & 0.6563 & 0.3424 & 75.94 \\
    + (a) + (b) + (c) \textbf{(\ourmodel)} & 18.08 & 0.6533 & 0.3277 & 63.77 \\
    + (a) + (b) + (c) + (d) \textbf{(\ourmodelplus)} & \textbf{18.32} & \textbf{0.6623} & \textbf{0.2789} & \textbf{49.44} \\
    \bottomrule
  \end{tabular}
  }
  \caption{\footnotesize \textbf{Ablation study of \ourmodelplus on Nerfbusters dataset}. We compare a Nerfacto baseline to: (a) directly running \ourmodeldm on rendered views without 3D updates, (b) distilling \ourmodeldm outputs via 3D updates in a non-incremental manner, (c) applying the 3D updates incrementally, and (d) add \ourmodeldm as a post-rendering step.}
  \vspace{-2mm}
  \label{tab:ablation_stages}
\end{table}

\vspace{-3mm}
\paragraph{Results.} Similar to \cref{sec:eval-in-the-wild}, our method outperforms its baselines across all metrics (\cref{tab:av_all_results}). \cref{fig:nre} illustrates how our method reduces artifacts across views in a consistent manner.

\subsection{Diagnostics}
\label{sec:ablations}

\paragraph{Pipeline components.} We ablate our method by applying our pipeline components incrementally. We compare a Nerfacto baseline to: (a) directly running \ourmodeldm on rendered views without 3D updates, (b) distilling \ourmodeldm outputs via 3D updates in a non-incremental manner, (c) applying the 3D updates incrementally, and (d) add \ourmodeldm as a post-rendering step. We show quantitative results in \cref{tab:ablation_stages} averaged over the Nerfbusters~\cite{warburg2023nerfbusters} dataset. Qualitative ablation can be found in \cref{fig:visual_ablation} and \cref{fig:compare_plus}. Simply applying \ourmodeldm to rendered outputs improves quality for renderings close to reference views but performs poorly in less observed regions, and causes flickering across rendered. Distilling diffusion outputs via 3D updates improves quality significantly but our incremental update strategy is essential, as evidenced by the degradation in LPIPS and FID when pseudo-views are added all at once.  Visualization of post-rendering results is provided in \cref{fig:compare_plus}, showcasing noticeable improvements in our outputs. These enhancements are further validated by the metric improvements shown in the last row of \cref{tab:ablation_stages}.

\begin{table}[t!]
  
  \centering
  \resizebox{\linewidth}{!}{
  \begin{tabular}{lcccccc}
    \toprule
    Method & $\tau$ & SD Turbo Pretrain. & Gram &  Ref & LPIPS$\downarrow$ & FID$\downarrow$ \\
    \midrule
    pix2pix-Turbo & 1000 & \checkmark & & & 0.3810 & 108.86  \\
    \midrule
    \ourmodeldm & 200 & \checkmark & & & 0.3190 & 61.80 \\
    \ourmodeldm & 200 & \checkmark & \checkmark &  & 0.3064 & 55.45  \\
    \ourmodeldm & 200 & \checkmark & \checkmark  & \checkmark & \textbf{0.2996} & \textbf{47.87}  \\
    \bottomrule
  \end{tabular}
  }
  \vspace{-2mm}
  \caption{\textbf{Ablation study of \ourmodeldm components on Nerfbusters dataset}. Reducing the noise level, conditioning on reference views, and incorporating Gram loss improve our model. }
  \label{tab:ablation}
  \vspace{-3mm}
\end{table}

\vspace{-3mm}
\paragraph{\ourmodeldm training.} We validate our \ourmodeldm training strategy by comparing to pix2pix-Turbo~\cite{parmar2024one}, which uses the same SD-Turbo backbone with a higher noise value ($\tau = 1000$ instead of $\tau = 200$) and to variants of our methods that omit reference view conditioning and Gram loss. \cref{tab:ablation} summarizes our results averages over the Nerfbusters dataset. Conditioning on reference views and with Gram loss further improves the result of our model. We note that simply decreasing the noise level from 1000 to 200 noticeably improves LPIPS and FID significantly, validating our findings in \cref{tab:psnr_ssim}. The primary reason is that high noise level causes the model to generate more hallucinated pixels that contradict the ground truth, resulting in poorer generalization on the test dataset. \textit{See Supp. Material for visual examples.}  

%% file: sec/6_conclusion.tex
\section{Conclusion}

We introduced \ourmodelplus, a novel pipeline for enhancing 3D reconstruction and novel-view synthesis. At its core is \ourmodeldm, a single-step diffusion model that can operate at near real time on modern NVIDIA GPUs. \ourmodeldm improves 3D representation quality through a progressive 3D update scheme and enables real-time artifact removal during inference. Compatible with both NeRF and 3DGS, it achieves a 2$\times$ improvement in FID scores over baselines while maintaining 3D consistency, showcasing its effectiveness in addressing artifacts and enhancing photorealistic rendering.

%% file: sec/X_suppl.tex
\clearpage
{
    \newpage
    \twocolumn[
        \centering
        \Large
        \textbf{Supplementary Material}\\
        \vspace{0.5em}
        \vspace{1.0em}
    ]
}

\renewcommand{\thefigure}{S\arabic{figure}}
\setcounter{figure}{0}
\renewcommand{\thetable}{S\arabic{table}}
\setcounter{table}{0}
\renewcommand{\thesection}{\Alph{section}}
\setcounter{section}{0}

We provide additional implementation details in \cref{sec:implementation_details} and further results in \cref{sec:additional_results}. We discuss limitations and future work in \cref{sec:limitations}.

\section{Additional Implementation Details}
\label{sec:implementation_details}
\subsection{Loss Functions}

We supervise our diffusion model with losses derived from readily available 2D supervision 
in the RGB image space, avoiding the need for any sort of 3D supervision that is hard to obtain:
\begin{itemize}
    \item \emph{Reconstruction loss.} Which we define as the L2 loss between the model output $\hat{I}$ and the ground-truth image $I$:
        \begin{equation}
        \mathcal{L}_{\text{Recon}} = \| \hat{I} - I \|_2.
        \end{equation}
    \item \emph{Perceptual loss.} We incorporate an LPIPS~\cite{johnson2016perceptual} loss based on the L1 norm of the VGG-16 features $\phi_l(\cdot)$ to enhance image details, defined as:
\begin{equation}
\mathcal{L}_{\text{LPIPS}} = \frac{1}{L} \sum_{l=1}^L \alpha_l \left\| \phi_l(\hat{I}) - \phi_l(I) \right\|_1,
\end{equation}
\item \emph{Style loss.}
    We use the Gram matrix loss based on VGG-16 features~\cite{reda2022film} to obtain sharper details. We define the loss as the L2 norm of the auto-correlation of VGG-16 features~\cite{reda2022film}:
\begin{equation}
\mathcal{L}_{\text{Gram}} = \frac{1}{L} \sum_{l=1}^L \beta_l \left\| G_l(\hat{I}) - G_l(I) \right\|_2,
\end{equation}
\noindent with the Gram matrix at layer $l$ defined as:
\begin{equation}
G_l(I) = \phi_l(I)^\top \phi_l(I).
\end{equation}
\end{itemize}
The final loss used to train our model is the weighted sum of the above terms: $\mathcal{L} = \mathcal{L}_{\text{Recon}} + \mathcal{L}_{\text{LPIPS}} + 0.5\mathcal{L}_{\text{Gram}}$.

\subsection{Progressive 3D updates}
Please refer to the pseudocode in \cref{alg:progressive_3d_updates} for further details.

\begin{algorithm}[t]
\caption{Progressive 3D Updates for Novel View Rendering}
\label{alg:progressive_3d_updates}
\KwIn{Reference views \( V_{\text{ref}} \), Target views \( V_{\text{target}} \), 
3D representation \( R \) (e.g., NeRF, 3DGS), Diffusion model \( D \) (\ourmodeldm),
Number of iterations per refinement \( N_{\text{iter}} \), Perturbation step size \( \Delta_{\text{pose}} \)}
\KwOut{High-quality, artifact-free renderings at \( V_{\text{target}} \)}

\textbf{Initialize:} Optimize 3D representation \( R \) using \( V_{\text{ref}} \).

\While{not converged}{
    \tcc{Optimize the 3D representation}
    \For{\( i = 1 \) to \( N_{\text{iter}} \)}{
        Optimize \( R \) using the current training set.
    }

    \tcc{Generate novel views by perturbing camera poses}
    \For{\textbf{each} \( v \in V_{\text{target}} \)}{
        Find the nearest camera pose of \( v \) in the training set.\\
        Perturb the nearest camera pose by \( \Delta_{\text{pose}} \).\\
        Render novel view \( \hat{v} \) using \( R \).\\
        Refine \( \hat{v} \) using diffusion model \( D \).\\
        Add refined view \( \hat{v} \) to the training set.
    }
}

\Return Refined renderings at \( V_{\text{target}} \).
\end{algorithm}

\begin{figure*}[t]
    \centering
    \includegraphics[width=\linewidth]{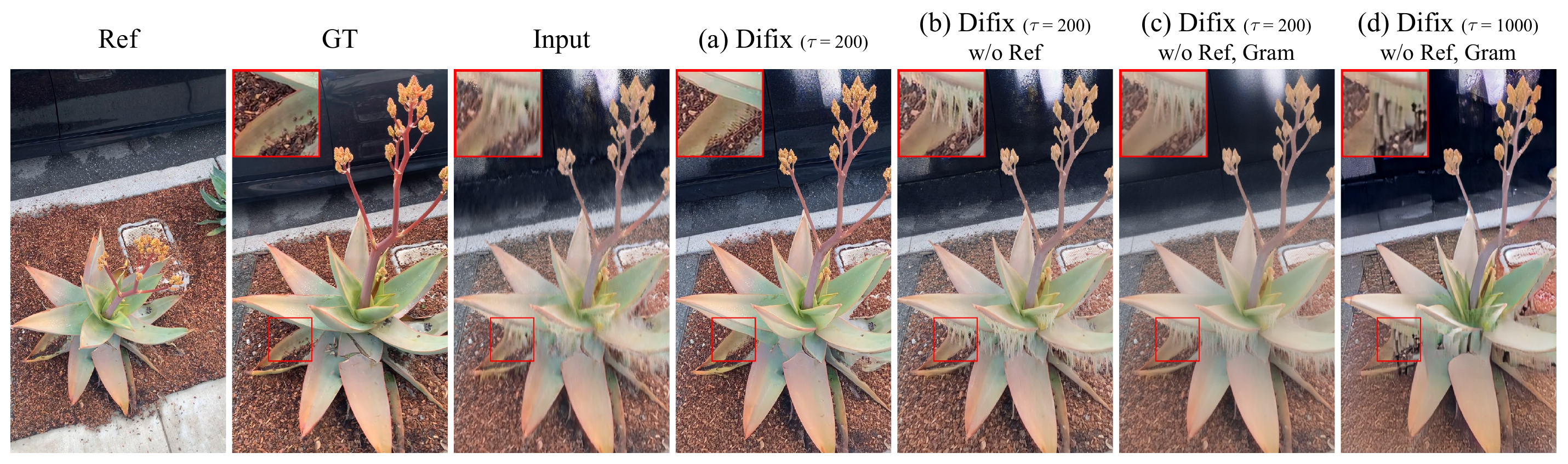}
    \caption{\footnotesize \textbf{Visual comparison of \ourmodeldm components.} Reducing the noise level $\tau$ ((c) \textit{vs.} (d)), incorporating Gram loss ((b) \textit{vs.} (c)), and conditioning on reference views ((a) \textit{vs.} (b)) all improve our model.} 
    \label{fig:ablation_fixer}
    \vspace{-2mm}
\end{figure*}

\subsection{Evaluation Metrics}

We employ several evaluation metrics to quantitatively assess the model’s performance in novel view synthesis. These metrics include Peak Signal-to-Noise Ratio (PSNR), Structural Similarity Index (SSIM)~\cite{wang2004image}, Learned Perceptual Image Patch Similarity (LPIPS)~\cite{johnson2016perceptual}, and Fréchet Inception Distance (FID)~\cite{heusel2017gans}. Following the evaluation procedure outlined by Nerfbusters~\cite{warburg2023nerfbusters}, we calculate a visibility map and mask out the invisible regions when computing the metrics.

\paragraph{PSNR.} 
The Peak Signal-to-Noise Ratio (PSNR) is widely used to measure the quality of reconstructed images by comparing them to ground truth images. It is defined as:
\begin{equation}
\text{PSNR} = 10 \cdot \log_{10} \left( \frac{\text{MAX}^2}{\text{MSE}} \right),
\end{equation}
where \(\text{MAX}\) represents the maximum possible pixel value (e.g., 255 for 8-bit images), and \(\text{MSE}\) is the mean squared error between the predicted image \(I_\text{pred}\) and the ground truth image \(I_\text{gt}\). Higher PSNR values indicate better reconstruction quality.

\paragraph{SSIM.}
The Structural Similarity Index (SSIM) evaluates the perceptual similarity between two images by considering luminance, contrast, and structure. It is computed as:
\begin{equation}
\text{SSIM}(I_\text{pred}, I_\text{gt}) = \frac{(2 \mu_\text{pred} \mu_\text{gt} + C_1)(2 \sigma_{\text{pred,gt}} + C_2)}{(\mu_\text{pred}^2 + \mu_\text{gt}^2 + C_1)(\sigma_\text{pred}^2 + \sigma_\text{gt}^2 + C_2)},
\end{equation}
where \(\mu\) and \(\sigma^2\) represent the mean and variance of the pixel intensities, respectively, and \(\sigma_{\text{pred,gt}}\) is the covariance. The constants \(C_1\) and \(C_2\) stabilize the division to avoid numerical instability.

\paragraph{LPIPS.}
The Learned Perceptual Image Patch Similarity (LPIPS) metric evaluates the perceptual similarity between two images based on feature embeddings extracted from pre-trained neural networks. It is defined as:
\begin{equation}
\text{LPIPS}(I_\text{pred}, I_\text{gt}) = \sum_{l} \| \phi_l(I_\text{pred}) - \phi_l(I_\text{gt}) \|_2^2,
\end{equation}
where \(\phi_l\) represents the feature maps from the \(l\)-th layer of a pre-trained VGG-16 network~\cite{simonyan2014very}. Lower LPIPS values indicate greater perceptual similarity.

\paragraph{FID.}
The Fréchet Inception Distance (FID) measures the distributional similarity between generated images and real images in the feature space of a pre-trained Inception network. It is computed as:
\begin{equation}
\text{FID} = \| \mu_\text{gen} - \mu_\text{real} \|_2^2 + \text{Tr}(\Sigma_\text{gen} + \Sigma_\text{real} - 2 (\Sigma_\text{gen} \Sigma_\text{real})^{\frac{1}{2}}),
\end{equation}
where \((\mu_\text{gen}, \Sigma_\text{gen})\) and \((\mu_\text{real}, \Sigma_\text{real})\) denote the means and covariances of the feature distributions for the generated and real images, respectively. Lower FID values indicate better alignment between the generated and real image distributions. We report the FID score calculated between the novel view renderings and the corresponding ground-truth images across the entire testing set.

\begin{figure*}[t]
    \centering
    \includegraphics[width=0.88\linewidth]{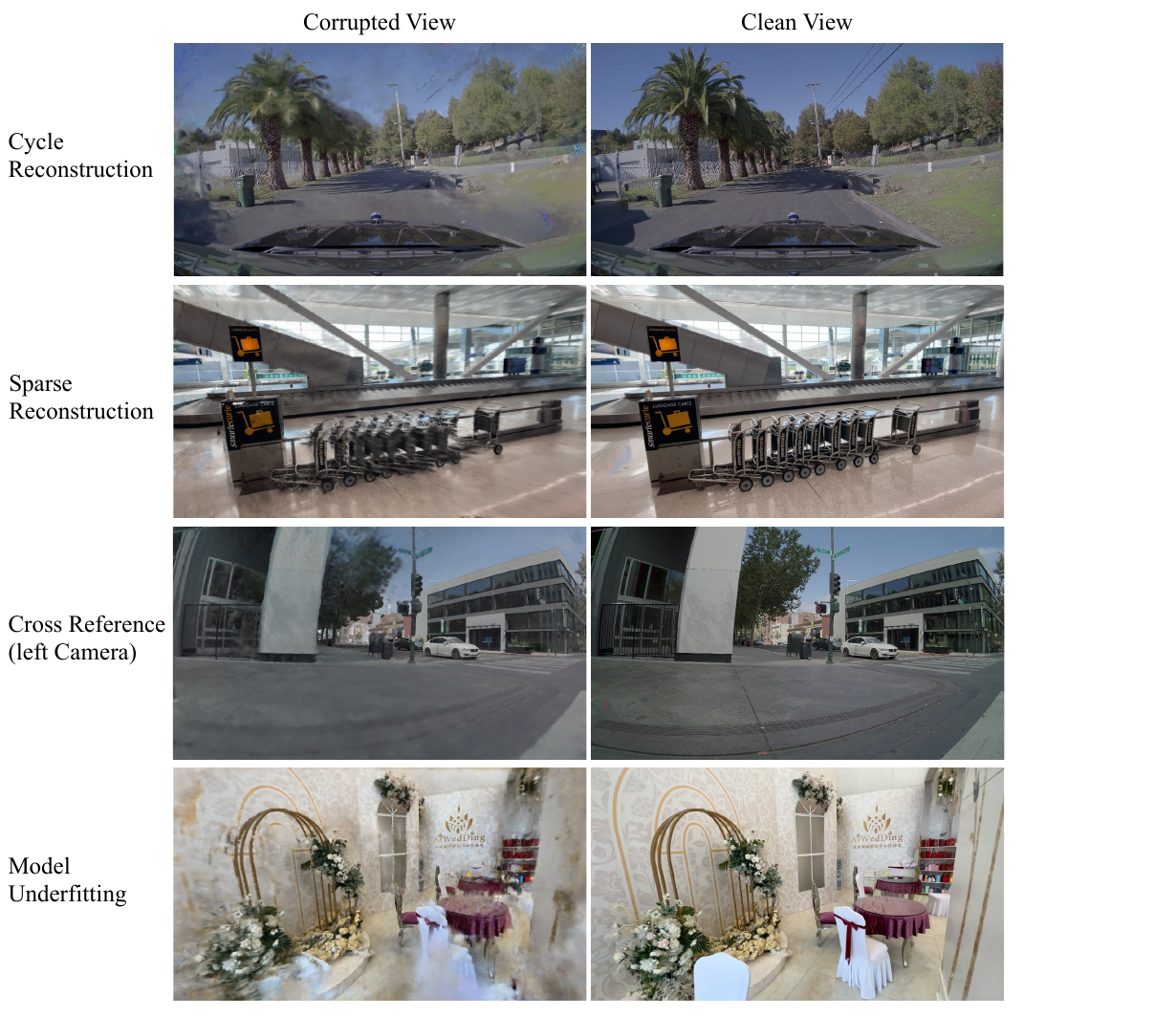}
    \caption{\footnotesize \textbf{Visualization of the paired dataset:}  We utilize a variety of strategies to simulate corrupted training data, including sparse reconstruction, cycle reconstruction, cross-referencing, and intentional model underfitting. The curated paired dataset provides a strong learning signal for the \ourmodeldm model.} 
    \label{fig:data_curation}
\end{figure*}

\subsection{Data Curation}

To curate paired training data, we employ a range of strategies including sparse reconstruction, cycle reconstruction, cross-referencing, and intentional model underfitting. The curated paired data generated through these strategies is visualized in \cref{fig:data_curation}. The simulated corrupted images exhibit common artifacts observed in extreme novel views, such as blurred details, missing regions, ghosting structures, and spurious geometry. This curated dataset provides a robust learning signal for the \ourmodeldm model, enabling the model to effectively correct artifacts in underconstrained novel views and enhance the quality of 3D reconstruction.

\section{Additional Results}
\label{sec:additional_results}

\subsection{Ablation Study of {\large \ourmodeldm}}
In addition to the quantitative results presented in \cref{tab:ablation}, we provide visual examples in \cref{fig:ablation_fixer} to demonstrate the effectiveness of our key design choices in \ourmodeldm. Compared to using a high noise level (\eg, pix2pix-Turbo~\cite{parmar2024one}), reducing the noise level significantly removes artifacts and improves overall visual quality ((c) \textit{vs.} (d)). Incorporating Gram loss enhances fine details and sharpens the image ((b) \textit{vs.} (c)). Furthermore, conditioning on a reference view corrects structural inaccuracies and alleviates color shifts ((a) \textit{vs.} (b)). Together, these advancements culminate in the superior results achieved by \ourmodeldm.

\subsection{Evaluation of Multi-View Consistency} 
We evaluate our model using the Thresholded Symmetric Epipolar Distance (TSED) metric~\cite{yu2023long}, which quantifies the number of consistent frame pairs in a sequence. As shown in Tab.~\ref{tab:tsed}, our model achieves higher TSED scores than reconstruction-based methods (\eg, Nerfacto) and other baselines, demonstrating superior multi-view consistency in novel view synthesis. Notably, the final post-processing step (\ourmodelplus) enhances image sharpness without compromising 3D coherence.

\begin{table}[h!]
    \centering
    \resizebox{\linewidth}{!}{
    \begin{tabular}{@{}l|cccccc@{}}
    \hline
    \textbf{Method} & Nerfacto & NeRFLiX & GANeRF & \ourmodel & \ourmodelplus \\ \hline
    \textbf{TSED ($T_{error}=2$)} & 0.2492 & 0.2532 & 0.2399 & 0.2601 & \textbf{0.2654} \\
    \textbf{TSED ($T_{error}=4$)} & 0.5318 & 0.5276 & 0.5140 & 0.5462 & \textbf{0.5515} \\ 
    \textbf{TSED ($T_{error}=8$)} & 0.7865 & 0.7789 & 0.7844 & \textbf{0.7924} & 0.7880 \\ \hline
    \end{tabular}
    }
    \caption{\footnotesize \textbf{Multi-view consistency evaluation on the DL3DV dataset.} A higher TSED score indicates better multi-view consistency.}
    \label{tab:tsed}
\end{table}

\section{Limitation and Future Work}
\label{sec:limitations}
We present \ourmodelplus, a novel pipeline designed to advance 3D reconstruction and novel-view synthesis. However, as a 3D enhancement model, the performance of \ourmodelplus is inherently limited by the quality of the initial 3D reconstruction. It currently struggles to enhance views where 3D reconstruction has entirely failed. Addressing this limitation through the integration of modern diffusion model priors represents an exciting direction for future research. To prioritize speed and approach near real-time post-rendering processing, \ourmodeldm is derived from a single-step image diffusion model. Additional promising avenues include scaling \ourmodeldm to a single-step video diffusion model, enabling enhanced long-context 3D consistency.